\documentclass[conference]{IEEEtran}
\IEEEoverridecommandlockouts
\usepackage{cite}
\usepackage{enumitem}
\usepackage{amsmath,amssymb,amsfonts}
\usepackage{algorithmic}
\usepackage{graphicx}
\usepackage{textcomp}
\usepackage{xcolor}
\usepackage{booktabs} 
\usepackage{amsmath} 
\usepackage{siunitx}
\usepackage{subcaption}
\usepackage{caption}
\usepackage{makecell}
\DeclareCaptionFont{eightpt}{\fontsize{8}{9.6}\selectfont}   
\def\BibTeX{{\rm B\kern-.05em{\sc i\kern-.025em b}\kern-.08em
    T\kern-.1667em\lower.7ex\hbox{E}\kern-.125emX}}
\begin{document}

\title{SIFPBPNet: A Dual-Path Network for Wearable and Cuffless Blood Pressure Estimation via Individualized Steady-state Representation
}

\author{Shuailong Tang$^{1}$, Xiaoyu Li$^{2}$, Donglin Xie$^{3}$, Wei Chen$^{4}$, Guangpu Zhu$^{2}$, Yelei Li$^{2}$, Yali Zheng$^{*1}$\\
$^{1}$ Shenzhen Technology University, Shenzhen, China\\
$^{2}$ OPPO Health Lab, Shenzhen, China \\
$^{3}$ Peking University, Beijing, China \\
$^{4}$ Southeast University, Nanjing, China
\thanks{Accepted for publication at the 48th Annual International Conference of the IEEE Engineering in Medicine and Biology Society (EMBC 2026), Toronto, Canada. This work was supported by Young Scientists Fund from National Natural Science Foundation of China (NSFC) (62301333). *Corresponding author: zhengyali@sztu.edu.cn. \copyright\ 2026 IEEE. Personal use of this material is permitted. Permission from IEEE must be obtained for all other uses, in any current or future media, including reprinting/republishing this material for advertising or promotional purposes, creating new collective works, for resale or redistribution to servers or lists, or reuse of any copyrighted component of this work in other works.}
}

\maketitle

\begin{abstract}
Continuous and cuffless blood pressure (BP) monitoring using photoplethysmography (PPG) is of great interest for low-cost and personalized cardiovascular health management. However, significant population heterogeneity and the ``one-to-many mapping'' problem---where similar waveforms across individuals correspond to different BP levels---limit the accuracy of conventional population-based models. To address this challenge, we propose a dual-path architecture termed SIFPBPNet, which separately represents steady-state and instantaneous features, through a Steady-state Feature Path (SFP) and an Instantaneous Feature Path (IFP). The SFP employs a Graph Attention Network (GAT) to extract individual-specific and long-term characteristics from multi-day historical PPG trajectories. In parallel, the IFP captures short-term dynamics from current PPG segments and incorporates the steady-state prior via a cross-attention mechanism. Experiments on a large-scale wearable dataset demonstrate that, SIFPBPNet achieves a Mean Absolute Error (MAE) of 8.57 and 5.97 mmHg for systolic and diastolic BP, respectively, outperforming state-of-the-art models. Furthermore, the SFP module consistently improves performance when integrated into various backbone architectures, yielding 2.8-13.1\% relative MAE reductions for systolic BP. These results highlight the strong generalizability and plug-and-play transferability of the SFP module, underscoring its great potential for accurate cuffless BP monitoring.

\end{abstract}

\begin{IEEEkeywords}
Cuffless blood pressure estimation, Photoplethysmography (PPG), Graph neural networks, Individualized representation learning.

\end{IEEEkeywords}

\section{Introduction}
Hypertension is one of the prevalent yet modifiable cardiovascular risk factor worldwide, which imposes a substantial burden on public health. According to the World Health Organization's 2023 Global Report on Hypertension, more than one billion adults are affected by this condition, accounting for approximately $19\%$ of global mortality \cite{kario2024} \cite{islam2022}. While traditional cuff-based  blood pressure (BP) measurement remains the gold standard for diagnosis, its intermittent measurement nature may lead to misdiagnosis, such as white-coat hypertension, masked hypertension and nocturnal hypertension. Moreover, the discomfort associated with repeated cuff inflation limits patient adherence and hinders its suitability for proactive BP management. Therefore, there is a great demand for unobtrusive solutions to enable precise risk management \cite{kario2020}.

Wearable and cuffless BP monitoring, particularly through Photoplethysmography (PPG) technique, has attracted substantial research interest in recent years. Most of existing deep learning approaches adopt a population-based training strategy (i.e., general models)\cite{liu2019}; however, they face a fundamental challenge, i.e., the ``one-to-many mapping'' problem. Recent research reveals that, up to $33\%$ of PPG segments with highly similar waveforms morphologies correspond to markedly different BP levels across individuals \cite{zheng2024}. As a result, these general models neglect individual-specific context and tend to regress toward the population mean, leading to degraded accuracy and limited generalization capability\cite{mieloszyk2022}. 

Recognizing the limitations of conventional population-based general models, it is necessary to develop an individual-aware modeling framework that can effectively represent individual physiological steady-state and incorporate it as individual prior knowledge. This transforms the BP estimation problem from pure ``instantaneous mapping'' to ``individual-prior estimation''. Based on this idea, we specifically propose the \textbf{Network of Steady-state and Instantaneous Feature Paths for BP (SIFPBPNet)}, which consists of dual feature extraction paths: \textbf{the Steady-state Feature Path (SFP)} for individual representation from long-term and historical PPG trajectories, and \textbf{the Instantaneous Feature Path (IFP)} for BP regression from instantaneous PPG segment by incorporating the extracted individual prior into the estimation network.

\section{Related Work}
The development of cuffless BP estimation models has transitioned through three major technological stages:
\subsection{Physiological-driven Stage}
Early research focused on using physiological indicators for BP estimation based on either explicit physiological laws or machine learning algorithms. Chen et al. \cite{chen2000} estimated high-frequency components of systolic BP (SBP) from pulse transit time (PTT). Poon et al. \cite{poon2005} established an inverse relationship between BP and the square of PTT. Subsequent pulse wave analysis methods utilized multiple morphological PPG features to correlate with hemodynamic variables such as peripheral resistance \cite{lin2020}. To improve robustness, Yang et al. \cite{yang2021} and Haddad et al. \cite{haddad2022} fused PPG morphological features with PTT, and employed random forests and linear regression for generalized BP estimation.
\subsection{Deep Learning-driven Stage}
To overcome the limitations of hand-crafted feature engineering, researchers utilized deep neural networks for more informative signal representations. Liu et al. \cite{liu2019} employed VGGNet to extract local time-domain features from pulse pressure wave signals. Ma et al. \cite{ma2024} designed a PPG Morphological Feature Learning (PMFL) algorithm to extract handcrafted features as prior knowledge to guide the SEM-ResNet model. Despite these advancements, significant estimation bias remains across individuals of different demographics \cite{theglobalcardiovascularriskconsortium2023}. Zheng et al. \cite{zheng2024} explicitly revealed the "one-to-many mapping" problem, where similar PPG waveforms correspond to substantially different BP levels, accounting for up to $33\%$ of cases.
\subsection{Personalized Modeling Stage}
The latest research integrates individual physiological backgrounds to alleviate the population heterogeneity issue. Zheng et al. \cite{zheng2024} used age, gender, and BMI as multi-modal attention query vectors to dynamically adjust waveform focus. Ma et al. \cite{ma2024} performed coarse classification of BP intervals followed by independent regression within subspaces. Zhang et al. \cite{zhang2022} employed disease labels for pre-stratification via random forests and trained dedicated BiLSTM models. 

Unlike previous methods that rely on static demographic attributes, our proposed SIFPBPNet introduces an explicit pathway to learn individualized steady-state representations from long-term (10-day) historical PPG trajectories. This approach provides a dynamic, personalized prior to address the challenges of existing population-based models when applied to highly heterogeneous populations.  

\section{Methodology}

\begin{figure*}[t]
  \centering
  \includegraphics[width=\linewidth]{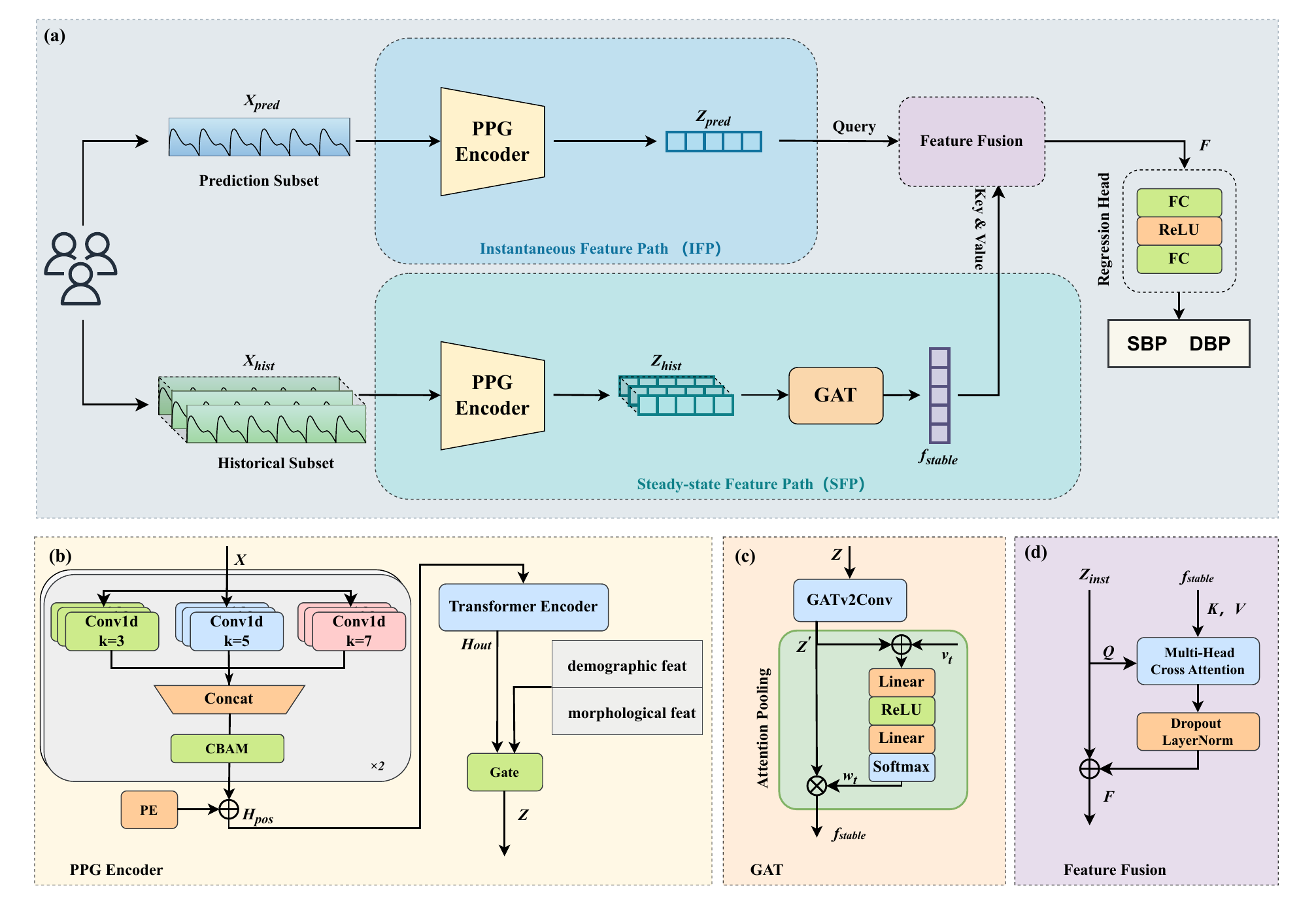}
  \vspace{-0.7cm}
  \caption{(a) The architecture of SIFPBPNet, comprising dual paths---SFP and IFP-and a Feature Fusion module. (b) Structure of the PPG Encoder, which extracts PPG latent features for both SFP and IFP. (c) Architecture of the GAT, employed to mine BP-related steady-state features from a sequence of historical PPG latent representations. (d) Design of the Feature Fusion module, integrating the dual-path features into a unified representation.}
  \vspace{-0.1cm}
  \label{fig:method:overview}
\end{figure*}

\subsection{Model Overview}

As shown in Fig.~\ref{fig:method:overview}(a), SIFPBPNet adopts a dual-path architecture, which extracts instantaneous (short-term) and steady-state (long-term) features related to individual BP status, respectively, achieving personalized BP estimation. SIFPBPNet consists of the two paths:

\begin{itemize}
        \item \textbf{IFP:} This path extracts instantaneous features from the current PPG record via a PPG encoder, fuses them with morphological and demographic features, and injects the individual long-term information extracted by the SFP path via a gated fusion mechanism to provide personalized constraints for instantaneous BP regression. 
    
    \item \textbf{SFP:} This path extracts steady-state features from historical PPG trajectories of each individual over multiple days via a Graph Neural Network (GNN). It is pre-trained with various BP-related tasks to extract individual steady-state prior information.
\end{itemize}

SIFPBPNet adopts a two-stage training strategy: the SFP path is pre-trained using historical data, which were fixed and then followed by the training of the IFP path. 

\subsection{PPG Encoder}

SFP and IFP share the same PPG encoder design to extract deep representations from PPG signals in their respective paths. The input of the encoder is a three-channel signal $\mathbf{X}$: the raw PPG signal, its first and second derivatives, and $L$ represents the length of the signal. 

As shown in Fig.~\ref{fig:method:overview}(b), the encoder integrates a Multi-Receptive Field convolutional neural network (MRF-CNN) to extract multi-scale local features and a Transformer to capture long-range dependencies. This combination yields comprehensive deep representations for both SFP and IFP.

First, the MRF-CNN captures features at different scales through parallel multi-scale convolutions with kernel sizes $k \in \{3, 5, 7\}$. For each kernel size $k$, the output feature $\mathbf{H}_k$ is calculated as:

\begin{equation}
    \mathbf{H}_k = \text{ReLU}(\text{BN}(\text{Conv1d}(\mathbf{X})))
\end{equation}

\noindent where $\mathbf{H}_k$ has $C_k$ channels and length $L'$ after convolution. 

The resulting multi-scale features are concatenated along the channel dimension and subsequently refined by a Convolutional Block Attention Module (CBAM). By employing the channel and spatial attention mechanisms, the CBAM highlights important features while suppressing noise, yielding the enhanced features $\mathbf{H_{\text{CBAM}}}$. 

To preserve temporal information, positional embeddings (PE) are injected into the sequence. The position information corresponds to the absolute timing of the PPG segments. For position $pos$ and dimension index $i$, the encoding is defined using sine and cosine functions: 

\begin{equation}
    PE_{(pos, 2i)} = \sin\left(\frac{pos}{10000^{2i/d_{\text{model}}}}\right)
\end{equation}

\begin{equation}
    PE_{(pos, 2i+1)} = \cos\left(\frac{pos}{10000^{2i/d_{\text{model}}}}\right)
\end{equation}
\noindent where $d_{\text{model}}$ denotes the model dimension. This positional encoding is added to the enhanced features to form the Transformer input:

\begin{equation}
    \mathbf{H}_{\text{pos}} = \mathbf{H}_{\text{CBAM}} + \mathbf{PE}
\end{equation}

Finally, the output feature vector for each PPG segment is obtained by applying global average pooling (GAP) of the transposed Transformer representations $\mathbf{H}_{\text{trans}}^T$:

\begin{equation}
    \mathbf{H}_{\text{out}} = \text{GAP}(\mathbf{H}_{\text{trans}}^T)
\end{equation} 

\noindent where $\mathbf{H}_{\text{out}}$ is a feature vector of dimension $C_{\text{feat}}$. 

\subsection{SFP}

Building upon the PPG encoder, the SFP aggregates deep features from a sequence of historical PPG records of a user to extract individualized steady-state information. In practical applications, the number of historical PPG records often varies between individuals, which poses challenges for traditional fixed-input methods to accommodate. 

To address this, we employ a Graph Attention Network (GAT), which adaptively aggregates information from a variable number of nodes, allowing the SFP to flexibly handle different amounts of historical records. The GAT is trained to learn BP-related signatures by optimizing for multiple downstream auxiliary tasks, including: age regression, individual-level hypertension detection, and high BP variability (BPV) detection. The structure of the GAT is shown in Fig.~\ref{fig:method:overview}(c). The details are described below.

 \emph{Node Feature Extraction}. First, node features are extracted using the PPG encoder. For a graph $G=(V, E)$, $V$ is the set of nodes and $E$ is the set of edges, and each node $V_i$ represents the latent representations $\mathbf{z}_i$ of a historical PPG segment obtained via the PPG encoder, as expressed as:

\begin{equation}
    \mathbf{z}_i = \text{PPGEncoder}(\mathbf{X}_i)
\end{equation}

\noindent The features of $N$ nodes for an individual are stacked to form the node feature matrix $\mathbf{Z}$ with $N$ rows and $C_{\text{feat}}$ columns, which serves as the input to the GAT. 

\emph{Graph Attention Convolution}. To effectively aggregate neighborhood information, the node features are processed through GATv2, an improved graph convolutional layer with attention mechanism. For single head attention, it calculates the attention weight $\alpha_{ij}$ between node $i$ and its neighbor node $j$ as expressed in Eq.~(7):

\begin{equation}
    \alpha_{ij} = \frac{\exp(\mathbf{a}^T \phi(\mathbf{W}[\mathbf{z}_i \| \mathbf{z}_j]))}{\sum_{k \in \mathcal{N}_i} \exp(\mathbf{a}^T \phi(\mathbf{W}[\mathbf{z}_i \| \mathbf{z}_k]))}
\end{equation}

\noindent where $\mathbf{W}$ is a weight matrix, $\mathbf{a}$ is an attention vector, $\phi$ denotes the LeakyReLU activation function, $\mathcal{N}_i$ denotes number of neighbor nodes, and $\|$ denotes concatenation. Then, the features from all $H$ attention heads are concatenated to form the final updated feature representation $\mathbf{z}_i'$ for node $i$, expressed in Eq.~(8):

\begin{equation}
    \mathbf{z}_i' = \Big\|_{h=1}^{H} \sigma\left(\sum_{j \in \mathcal{N}_i} \alpha_{ij}^{(h)} \mathbf{W}^{(h)} \mathbf{z}_j\right)
\end{equation}

\noindent where $\|$ denotes concatenation, $H$ is the number of attention heads, $h$ indexes the attention head, $\sigma$ denotes an activation function, and $\mathbf{W}^{(h)}$ is the weight matrix for head $h$.

\emph{Task-guided Attention Pooling}. To dynamically adjust node weights for specific tasks, we utilize task-guided attention pooling mechanism. For a given task $t$, the updated node feature $\mathbf{z_i}'$ is concatenated with a randomly initialized task embedding $\mathbf{v}_t$, and passed through a multi-layer perceptron (MLP) and normalized using a Softmax function to calculate the attention weight $w_i^{(t)}$ for each node: 

\begin{equation}
    w_i^{(t)} = \frac{\exp({\text{MLP}}([\mathbf{z}_i' \| \mathbf{v}_t]))}{\sum_{j=1}^{N} \exp({\text{MLP}}([\mathbf{z}_j' \| \mathbf{v}_t]))}
\end{equation}

Steady-state Feature Generation. Finally, the node features are weighted $w_i^{(t)}$, summed, and processed through another MLP to obtain the steady-state feature vector $f_{\text{stable}}$ in Eq.~(10). This vector carries the individualized steady-state prior for subsequent BP regression.

\begin{equation}
    f_{\text{stable}} = {\text{MLP}}\left(\sum_{i=1}^{N} w_i^{(t)} \mathbf{z}_i'\right)
\end{equation}

\subsection{IFP and Fusion Module}

The IFP is designed to estimate BP based on the current PPG segment, with individual prior from the SFP path. Initially, it extracts the instantaneous latent representations $\mathbf{z}_{\text{inst}}$ of the current signal by the shared PPG encoder. To enrich this representation, subject demographics $\mathbf{f}_{\text{demo}}$ and morphological features $\mathbf{f}_{\text{morph}}$ are fused with $\mathbf{z}_{\text{inst}}$ through a gated fusion mechanism. Finally, to achieve personalized BP estimation, the IFP incorporates the individualized steady-state prior extracted by the SFP via a cross-attention mechanism as shown in Fig.~\ref{fig:method:overview}(d). 

When SFP uses multi-task pre-training, the steady-state representation $\mathbf{f}$ is constructed by stacking the feature vectors from $T$ auxiliary tasks ($\mathbf{f}$ = $[\mathbf{f}_1; \mathbf{f}_2; \ldots; \mathbf{f}_T]$). The cross-attention mechanism operates by treating the instantaneous representations $\mathbf{z}_{\text{inst}}$ as the Query, while the steady-state feature $\mathbf{f}$ as the Key and Value. The specific projections are computed as follows:

\begin{equation}
    \mathbf{Q} = \mathbf{z}_{\text{inst}} \mathbf{W}_Q
\end{equation}

\begin{equation}
    \mathbf{K} = \mathbf{f} \mathbf{W}_K, \quad \mathbf{V} = \mathbf{f} \mathbf{W}_V
\end{equation}

\noindent where $\mathbf{W}_Q$, $\mathbf{W}_K$, and $\mathbf{W}_V$ are learnable weight matrices that project the features into a common dimension $d_k$. The final fused representation $\mathbf{F}$ is computed as:

\begin{equation}
    \mathbf{F} = \text{softmax}\left(\frac{\mathbf{Q} \mathbf{K}^T}{\sqrt{d_k}}\right) \mathbf{V}
\end{equation}

\noindent Subsequently, the unified feature vector $\mathbf{F}$ is fed into the regression head to generate the final BP estimation.

\subsection{Training Setup}

The training process of the entire framework is divided into two stages. In the SFP pre-training stage, historical data from 40\% of individuals were used. To enable SFP to learn steady-state information from historical data, three BP-related tasks were employed: age regression, individual-level hypertension detection according to Hypertension Management Guidelines, and individual-level high BP variability (BPV) detection. When the standard deviations of SBP or DBP over the historical period are beyond 10 or 7 mmHg, this individual is marked as high BPV.

The selection of these auxiliary tasks is grounded in their physiological relevance to BP regulation. Age regression captures vascular aging patterns, as PPG waveform morphology is known to correlate with arterial stiffness and age-related vascular changes. Hypertension detection and high BPV detection directly target BP-related phenotypes, leveraging the fact that historical PPG trajectories encode individual-specific hemodynamic signatures associated with chronic BP elevation and visit-to-visit BP variability. These three tasks collectively provide complementary steady-state priors: age reflects long-term vascular structural changes, while hypertension and BPV status characterize functional BP dynamics.

In the IFP training stage, data from the remaining 60\% of individuals were used for training and testing, ensuring data isolation between the two stages. For IFP training, the data were split at a ratio of 4:1:1 on an individual basis, ensuring that data from the same individual did not appear across different sets.

\section{Experiments}
\subsection{Dataset}
This study is conducted based on the OPPO HBPM dataset, a large-scale wearable BP dataset derived from previous research \cite{li2024}. During the data collection process, subjects were required to synchronously record their BP and PPG signals across three daily periods---morning, noon, and evening---over a duration of approximately one to two months. Concurrently, demographic characteristics were collected, and PPG morphological features were extracted post-acquisition. In this study, we selected 10 consecutive days of data from each individual, ensuring at least two time periods per day, to serve as historical data. Data recorded subsequent to these historical periods were utilized for model prediction.

\begin{figure}[htbp]
  \centering
  \includegraphics[width=\linewidth]{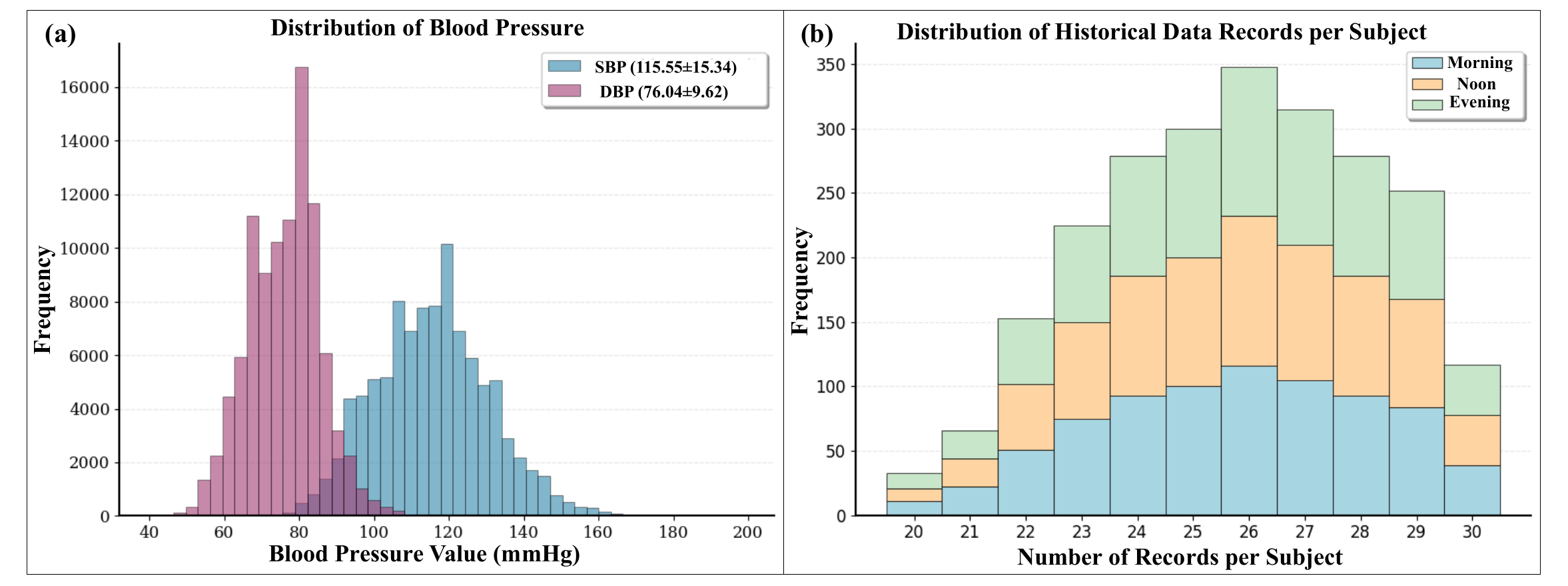}
  \caption{(a) Distribution and range of SBP and DBP of in the BP regression data. (b) Distribution of historical data records per subject}
    \vspace{-0.3cm}
  \label{fig:distribution}
\end{figure}

The distribution of the data is shown in Table~\ref{tab:dataset}. The dataset includes 790 individuals, with males accounting for 46.33\% and females for 53.67\%, and the age of $38.53 \pm 9.93$ years old. PPG signals were acquired for 60 seconds at a sampling rate of 250 Hz, with synchronized ground-truth BP values collected using an Omron cuff-based oscillometric BP monitor. After partitioning the data into historical and target prediction sets and excluding low-quality signals, we obtained 20,342 historical samples and 98,258 prediction samples. Fig.~\ref{fig:distribution}(a) illustrates the BP distribution of the prediction subset, while Fig.~\ref{fig:distribution}(b) shows the distribution of the number of historical data samples per individual.

\begin{table}[htbp]
\centering
\caption{Statistics of prediction and historical subsets}
\label{tab:dataset}
\begin{tabular}{lcc}
\toprule
\textbf{Parameter} & \textbf{Prediction subset} & \textbf{Historical subset} \\
\midrule
Number of subjects (M/F) & 790 (366/424) & 790 (366/424) \\
Age (years old) & $38.53\pm9.93$  & $38.53\pm9.93$  \\
Records (n)                  & 98\,258        & 20\,342        \\
SBP (mmHg)                & $115.55\pm15.34$ & $114.09\pm15.77$ \\
DBP (mmHg)                & $76.04\pm9.62$  & $75.74\pm10.16$ \\
\bottomrule
\end{tabular}
\end{table}

\subsection{Preprocessing}
The dataset underwent several preprocessing steps. First, a fourth-order Butterworth filter with a passband of 0.5 Hz to 12 Hz was applied to the raw PPG signals to remove baseline drift and high-frequency noise. Subsequently, a template-matching quality detection method was employed to filter out low-quality PPG signals and select high-quality PPG segments with a duration of 12 seconds and a length of 3,000 sampling points from the complete PPG waveforms.

\section{Results}
\subsection{BP Estimation Performance}
Table \ref{tab:overall_performance} lists the BP estimation performance of SIFPBPNet on the OPPO HBPM dataset. The Mean Absolute Errors (MAEs) for SBP and DBP are 8.57  and 5.98 mmHg, respectively. For SBP, the Mean Error $\pm$ Standard Deviation of Error (ME $\pm$ SDE) is $0.02 \pm 10.76$ mmHg, with the proportion of prediction errors $<5$ mmHg, $<10$ mmHg, and $<15$ mmHg being 35.83\%, 64.99\%, and 84.20\%, respectively. For DBP, the ME $\pm$ SDE is $-0.10 \pm 7.70$ mmHg, and the proportion of prediction errors $<5$ mmHg, $<10$ mmHg, and $<15$ mmHg is 51.67\%, 81.79\%, and 93.88\%, respectively.

\begin{table}[h]
\centering
\caption{Overall BP Estimation Performance of SIFPBPNet}
\label{tab:overall_performance}
\begin{tabular}{lcccc}
\toprule
 & \makecell{MAE\\(mmHg)} & $R^2$ & \makecell{ME$\pm$SDE\\(mmHg)} \\ \midrule
DBP & 5.98 & 0.34 & $-0.10\pm7.70$ \\
SBP & 8.57 & 0.44 & \phantom{$-$}0.02$\pm10.76$ \\ \bottomrule
\end{tabular}
\end{table}


To further validate the superiority of SIFPBPNet, we compared it against several advanced deep learning architectures, including ResNet18, VGGNet\cite{liu2019}, BiLSTM\cite{zhang2022} and SEM-ResNet\cite{ma2024}. For a fair end-to-end comparison, these baseline models are implemented as complete standalone architectures that take identical inputs (raw PPG, first and second derivatives, morphological features, and demographic features) and directly regress SBP and DBP values, without utilizing the SFP module or the dual-path fusion mechanism. As summarized in Table \ref{tab:comparison_models}, SIFPBPNet consistently outperformed all baseline models across all evaluation metrics. Notably, compared to the strongest baseline (SEM-ResNet for SBP and VGGNet for DBP), SIFPBPNet reduced the MAE of SBP by 8.5\% and DBP by 4.0\%. SIFPBPNet also demonstrates superior performance in the ME $\pm$ SDE metrics.

\begin{table}[h]
\centering
\scriptsize
\caption{Comparison with other deep learning models}
\label{tab:comparison_models}
\setlength{\tabcolsep}{0.8mm}
\renewcommand{\arraystretch}{1.05}

\begin{tabular}{lcccccc}
\toprule
\multicolumn{7}{c}{\textbf{DBP}} \\ \midrule
Model & \makecell{MAE\\(mmHg)} & $R^2$ & \makecell{ME$\pm$SDE\\(mmHg)} & \makecell{$<$5mmHg\\(\%)} & \makecell{$<$10mmHg\\(\%)} & \makecell{$<$15mmHg\\(\%)} \\ \midrule
ResNet18   & 6.65 & 0.22 & $1.30\pm8.24$ & 45.78 & 76.71 & 92.23 \\
VGGNet\cite{liu2019}     & 6.23 & 0.31 & $0.88\pm7.82$ & 48.52 & 79.52 & 93.99 \\
BiLSTM\cite{zhang2022}     & 6.62 & 0.21 & $1.31\pm8.29$ & 46.03 & 77.34 & 92.10 \\
SEM-ResNet\cite{ma2024} & 6.38 & 0.27 & $-0.91\pm8.00$ & 47.55 & 79.15 & 93.10 \\
\textbf{SIFPBP (Ours)} & \textbf{5.98} & \textbf{0.34} & $\mathbf{-0.10\pm7.70}$ & \textbf{51.67} & \textbf{81.79} & \textbf{93.88} \\ \bottomrule
\end{tabular}

\vspace{0.8ex}

\begin{tabular}{lcccccc}
\toprule
\multicolumn{7}{c}{\textbf{SBP}} \\ \midrule
Model & \makecell{MAE\\(mmHg)} & $R^2$ & \makecell{ME$\pm$SDE\\(mmHg)} & \makecell{$<$5mmHg\\(\%)} & \makecell{$<$10mmHg\\(\%)} & \makecell{$<$15mmHg\\(\%)} \\ \midrule
ResNet18   & 9.93 & 0.27 & $4.84\pm11.32$ & 30.59 & 57.01 & 77.32 \\
VGGNet\cite{liu2019}     & 10.24 & 0.24 & $1.13\pm12.54$ & 28.35 & 54.74 & 75.57 \\
BiLSTM\cite{zhang2022}     & 9.83 & 0.30 & $4.19\pm11.34$ & 30.03 & 57.16 & 77.73 \\
SEM-ResNet\cite{ma2024} & 9.36 & 0.34 & $1.47\pm11.61$ & 32.87 & 60.23 & 80.01 \\
\textbf{SIFPBP (Ours)} & \textbf{8.57} & \textbf{0.44} & $\mathbf{0.02\pm10.76}$ & \textbf{35.83} & \textbf{64.99} & \textbf{84.20} \\ \bottomrule
\end{tabular}
\end{table}

\subsection{Ablation Study}
To demonstrate the effectiveness of each module in SIFPBPNet and evaluate their contributions to model performance, we conducted ablation experiments on the SFP, Transformer, and CBAM modules, which are denoted as: (1) MRF-CNN only; (2) MRF-CNN + CBAM; (3) MRF-CNN + CBAM + Transformer; (4) The complete SIFPBPNet.

As shown in Table \ref{tab:ablation}, the progressive integration of CBAM and Transformer modules led to incremental improvements in MAE, suggesting that capturing both local spatial features and long-range temporal dependencies is crucial for BP modeling. However, the most significant performance leap occurred with the introduction of the SFP module. Specifically, the MAE for SBP dropped from 9.07 mmHg to 8.57 mmHg upon adding SFP. This result confirms that the steady-state physiological information extracted via SFP provides subject-specific context that conventional methods struggle to capture from current segments alone.

\begin{table}[h]
\centering
\scriptsize
\caption{Ablation Study Results}
\label{tab:ablation}
\setlength{\tabcolsep}{1.2mm}
\renewcommand{\arraystretch}{1.05}

\begin{tabular}{lcccc}
\toprule
 & \multicolumn{2}{c}{DBP} & \multicolumn{2}{c}{SBP} \\ \cmidrule(lr){2-3} \cmidrule(lr){4-5}
Configuration & \makecell{MAE\\(mmHg)} & \makecell{ME$\pm$SDE\\(mmHg)} & \makecell{MAE\\(mmHg)} & \makecell{ME$\pm$SDE\\(mmHg)} \\ \midrule
MRF-CNN & 6.31 & $-0.85\pm7.94$ & 9.34 & $-0.69\pm11.78$ \\
MRF-CNN+CBAM & 6.45 & $-1.60\pm8.02$ & 9.17 & $-0.87\pm11.52$ \\
MRF-CNN+CBAM+Trans. & 6.20 & $-0.20\pm7.90$ & 9.07 & $0.80\pm11.36$ \\
\textbf{SIFPBP} & \textbf{5.98} & $\mathbf{-0.10\pm7.70}$ & \textbf{8.57} & $\mathbf{0.02\pm10.76}$ \\ \bottomrule
\end{tabular}
\end{table}

\subsection{Results of SPF Pre-training Task Configurations}


The efficacy of the SFP module depends on the auxiliary tasks used during its pre-training phase. To investigate the contribution of each individual task, we first evaluated three single-task baselines: age regression only, hypertension detection only, and high BPV (HBPV) detection only. We then compared three multi-task combinations: (1) age regression + hypertension detection; (2) hypertension detection + HBPV detection; (3) age regression + hypertension detection + HBPV detection.

As shown in Table~\ref{tab:pretrain_tasks}, hypertension detection alone achieves the best single-task performance (SBP MAE: 8.61~mmHg, DBP MAE: 6.05~mmHg). Notably, adding age regression to hypertension detection maintains competitive SBP performance (8.57~mmHg) while yielding the best DBP result (5.98~mmHg), indicating that age-related physiological variations provide complementary information for diastolic pressure estimation.



\begin{table}[h]
\centering
\scriptsize
\caption{Impact of Different Pre-training Task Combinations}
\label{tab:pretrain_tasks}
\setlength{\tabcolsep}{1.2mm}
\renewcommand{\arraystretch}{1.05}

\begin{tabular}{lcccc}
\toprule
 & \multicolumn{2}{c}{DBP} & \multicolumn{2}{c}{SBP} \\ \cmidrule(lr){2-3} \cmidrule(lr){4-5}
Task Combination & \makecell{MAE\\(mmHg)} & \makecell{ME$\pm$SDE\\(mmHg)} & \makecell{MAE\\(mmHg)} & \makecell{ME$\pm$SDE\\(mmHg)} \\ \midrule
Age & 6.13 & $-0.98\pm7.73$ & 9.25 & $-1.70\pm11.51$ \\
Hyper & 6.05 & $1.13\pm7.47$ & 8.61 & $1.24\pm10.56$ \\
HBPV & 6.10 & $0.59\pm7.72$ & 9.34 & $1.54\pm11.58$ \\
Age + Hyper & \textbf{5.98} & $\mathbf{-0.10\pm7.70}$ & 8.57 & $0.02\pm10.76$ \\
Hyper + HBPV & 6.06 & $-0.74\pm7.74$ & \textbf{8.42} & $\mathbf{-0.30\pm10.70}$ \\
Age + Hyper + HBPV & 6.02 & $-0.26\pm7.72$ & 8.87 & $2.43\pm10.84$ \\ \bottomrule
\end{tabular}
\end{table}

\subsection{Transferability of SFP Module}

\begin{figure}[h]
  \centering
  \includegraphics[width=\columnwidth]{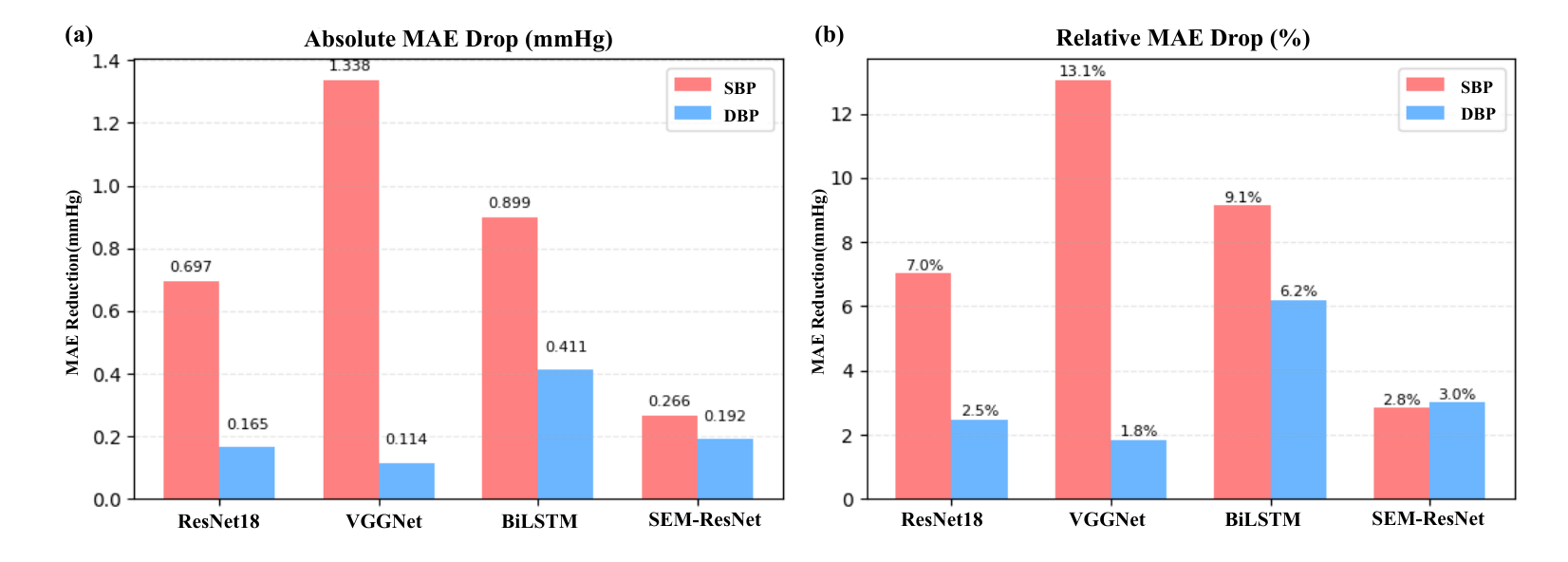}
  \vspace{-0.4cm}
  \caption{(a) Absolute reduction in MAE after integrating the SFP module. (b) Relative percentage reduction in MAE.}
  \vspace{-0.1cm}
  \label{fig:MAE}
\end{figure}

To evaluate the robustness and plug-and-play capability of the SFP module independently of the IFP design, we fixed the pre-training task to ``age regression + hypertension detection'' and integrated the SFP module with the baseline architectures by replacing the PPG Encoder within the IFP path while retaining the Feature Fusion module and regression head. Specifically, for each baseline model, its feature extraction layers serve as the PPG Encoder to generate instantaneous representations $\mathbf{z}_{\text{inst}}$, which are then fused with the steady-state prior $\mathbf{f}_{\text{stable}}$ from the SFP via the same cross-attention mechanism described in Sec.~III-D. As summarized in Table \ref{tab:SFP_comparison} and visualized in Fig. \ref{fig:MAE}, the addition of SFP leads to consistent performance enhancement across all tested models, especially for SBP.

\begin{table}[h]
\centering
\caption{Performance comparison with and without SFP module}
\label{tab:SFP_comparison}
\footnotesize
\setlength{\tabcolsep}{3pt}
\begin{tabular}{@{}lcccccc@{}}
\toprule
 & \multicolumn{3}{c}{DBP (mmHg)} & \multicolumn{3}{c}{SBP (mmHg)} \\ 
\cmidrule(lr){2-4} \cmidrule(lr){5-7}
Model & MAE & $R^2$ & ME$\pm$SDE & MAE & $R^2$ & ME$\pm$SDE \\ 
\midrule
\multicolumn{7}{@{}l}{\textbf{Without SFP:}} \\
ResNet18 & 6.65 & 0.22 & $1.30\pm8.24$ & 9.93 & 0.27 & $4.84\pm11.32$ \\
VGGNet & 6.23 & 0.31 & $0.88\pm7.82$ & 10.24 & 0.24 & $1.13\pm12.54$ \\
BiLSTM & 6.62 & 0.21 & $1.31\pm8.29$ & 9.83 & 0.30 & $4.19\pm11.34$ \\
SEM-ResNet & 6.38 & 0.27 & $-0.91\pm8.00$ & 9.36 & 0.34 & $1.47\pm11.61$ \\ 
\midrule
\multicolumn{7}{@{}l}{\textbf{With SFP:}} \\
ResNet18+SFP & 6.48 & 0.27 & $-0.66\pm8.03$ & 9.23 & 0.36 & $4.26\pm10.69$ \\
VGGNet+SFP & 6.11 & 0.31 & $-0.64\pm7.81$ & 8.91 & 0.39 & $0.91\pm11.23$ \\
BiLSTM+SFP & 6.21 & 0.31 & $1.00\pm7.81$ & 8.93 & 0.40 & $1.81\pm11.01$ \\
SEM-ResNet+SFP & 6.18 & 0.30 & $-0.07\pm7.90$ & 9.10 & 0.37 & $0.60\pm11.46$ \\ 
\bottomrule
\end{tabular}
\end{table}

Fig. \ref{fig:MAE}(a) illustrates the absolute reduction in MAE, while Fig. \ref{fig:MAE}(b) shows the relative percentage improvement. The SFP module exhibits a particularly powerful impact on SBP estimation. Notably, for the VGGNet backbone, the integration of SFP resulted in a substantial SBP MAE drop of 1.34 mmHg, equivalent to a 13.1\% relative reduction. Even for SEM-ResNet, which already performs well independently, the SFP module managed to further refine the predictions with a 2.8\% to 3.0\% improvement in SBP and DBP, respectively. This ubiquitous improvement signals that SFP distills an individualized and BP-related signature that complements any PPG feature extractor.

\section{Discussion \& Conclusion}
In this study, we address the "one-to-many mapping" problem inherent in conventional population-based modeling strategy in cuffless BP estimation by proposing the SIFPBPNet framework. The core innovation of SIFPBPNet lies in the introduction of the SFP pathway, which explicitly represents individual-specific, steady-state and BP-related information from long-term historical PPG data. Comprehensive evaluation on a large-scale, real-world wearable dataset demonstrates that, incorporating these steady-state features as an individual prior in the instantaneous BP regression stage, substantially enhanced estimation performance. The SFP module not only markedly reduces the MAE of SBP compared to the strongest baseline, but also exhibits excellent transferability across various deep learning backbones. These results suggest that, individual-level representation learning is a critical step toward reliable and personalized BP estimation. The proposed method may further serve as a generalizable strategy for other physiological signal modeling tasks affected by similar issues of intersubject variability.    


In conclusion, SIFPBPNet provides an effective and scalable framework for accurate daily BP measurement in real-world wearable scenarios involving heterogeneous populations. Future work will focus on three directions: (i) validating the proposed individual steady-state representation method in longitudinal study settings, with particular emphasis on its ability to track personalized baseline variations over time; (ii) systematically exploring additional auxiliary tasks and incorporating self-supervised and contrastive learning objectives into the SFP pre-training framework to reduce reliance on labeled data; (iii) evaluating varying PPG window lengths to optimize the trade-off between temporal context richness and practical deployment constraints.

\bibliographystyle{IEEEtran}
\bibliography{ref}

\end{document}